\documentclass[letterpaper]{article}

\usepackage{natbib,alifeconf}  %% The order is important
\usepackage{url,hyperref}
\usepackage{booktabs}

\usepackage{graphicx}
\usepackage{amsmath}
\usepackage{cleveref}
\usepackage{caption} % For two-column teaser

\usepackage{amssymb}
\usepackage{booktabs}
\usepackage{multirow}
\usepackage{multicol}
\usepackage[table]{xcolor}% http://ctan.org/pkg/xcolor

\usepackage[skip=6pt]{caption}

\usepackage{array}

\usepackage{pifont}% http://ctan.org/pkg/pifont
\definecolor{limegreen}{HTML}{32CD32}

\DeclareMathOperator*{\argmin}{arg\,min}

\newcommand{\rev}[1]{#1}

\title{NoiseNCA: Noisy Seed Improves Spatio-Temporal \\ Continuity of Neural Cellular
Automata}

\author{
    Ehsan Pajouheshgar,
    Yitao Xu,
    Sabine Süsstrunk \\
    \mbox{}\\
    School of Computer and Communication Sciences, EPFL, Switzerland \\
    {\tt\small \{ ehsan.pajouheshgar, yitao.xu, sabine.susstrunk \}@epfl.ch}
} % email of corresponding author

\begin{document}

\twocolumn[{%
\renewcommand\twocolumn[1][]{#1}%
\maketitle
\begin{center}
    \centering
    \captionsetup{type=figure}
    \vspace{-25pt}
    \includegraphics[width=\linewidth]{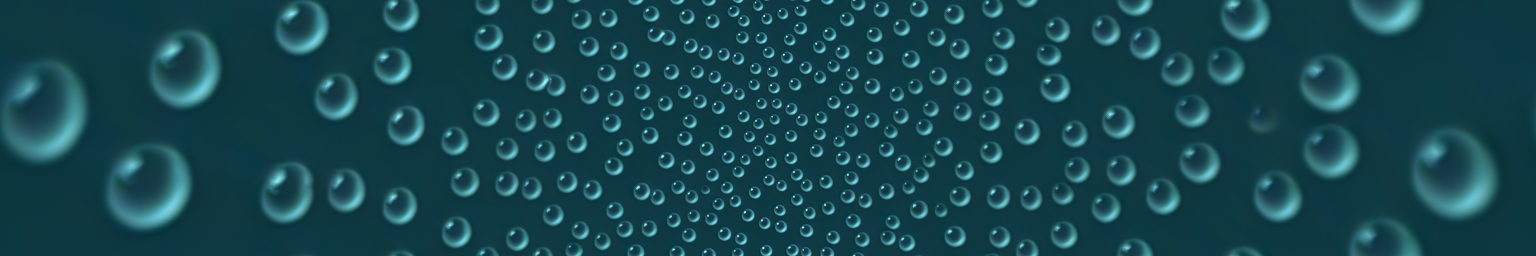}
    \includegraphics[width=\linewidth]{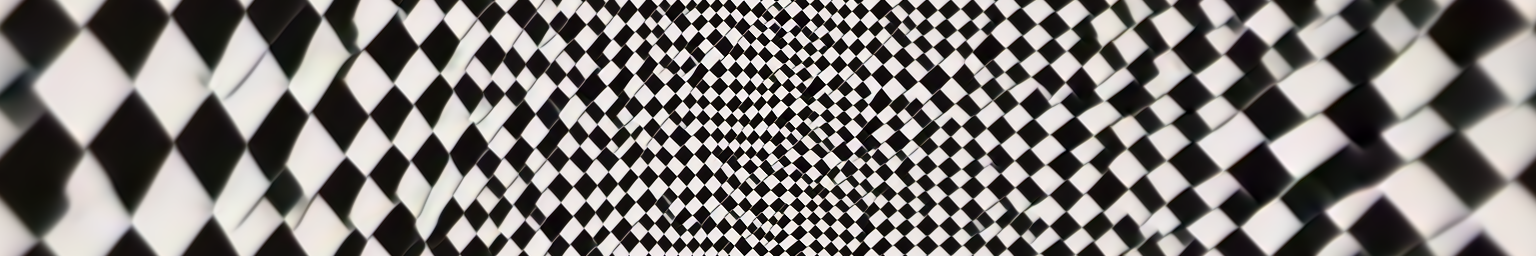}
    \includegraphics[width=\linewidth]{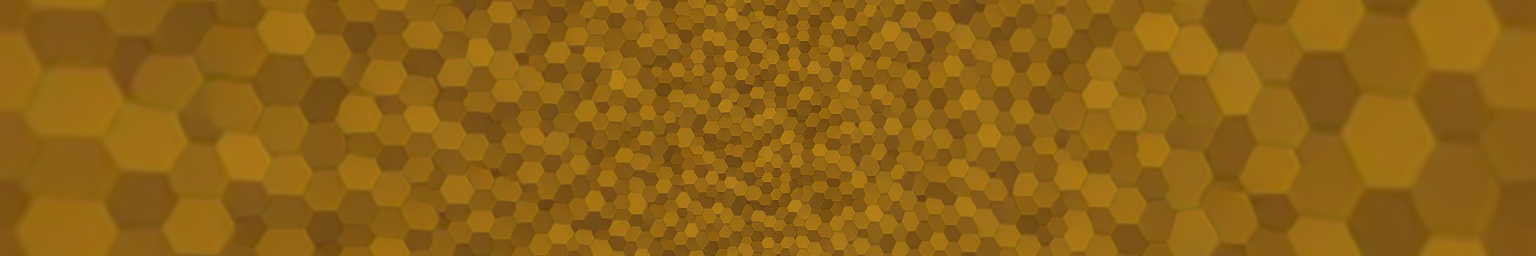}
    \captionof{figure}{The update rule learned by \textit{Noise-NCA} generalizes across different granularities of space and time discretization.
    This allows our model to synthesize textures at varying scales and speeds.
    Illustrated here, when changing the cell size across the space, \textit{Noise-NCA}
    creates textures at varying scales within the same grid. For more results and an interactive demo, please check our webpage at \url{https://noisenca.github.io}}
    \vspace{-1pt}
    \label{fig:teaser}
\end{center}%
}]

\begin{abstract}
    \vspace{-3pt}
    Neural Cellular Automata (NCA) is a class of Cellular Automata where the update rule is parameterized by a neural network that can be trained using gradient descent.
    In this paper, we focus on NCA models used for texture synthesis, where the update rule is inspired by partial differential equations (PDEs) describing reaction-diffusion systems. To train the NCA model, the \rev{spatio-temporal} domain is discretized, and Euler integration is used to numerically simulate the PDE. However, whether a trained NCA truly learns the continuous dynamic described by the corresponding PDE or merely overfits the discretization used in training remains an open question. We study NCA models at the limit where space-time discretization approaches continuity.
    We find that existing NCA models tend to overfit the training discretization, especially in the proximity of the initial condition, also called "seed". To address this, we propose a solution that utilizes uniform noise as the initial condition. We demonstrate the effectiveness of our approach in preserving the consistency of NCA dynamics across a wide range of spatio-temporal granularities. Our improved NCA model enables two new test-time interactions by allowing continuous control over the speed of pattern formation and the scale of the synthesized patterns. We demonstrate this new NCA feature in our interactive online demo. Our work reveals that NCA models can learn continuous dynamics and opens new venues for NCA research from a \rev{dynamical system's} perspective.
\end{abstract}

\vspace{-14pt}
\section{Introduction}
\vspace{-3pt}
For decades, Cellular Automata (CA) have served as the foundation for studying and creating life-like systems \citep{von-ca, ca-conway},
demonstrating how simple local interactions can lead to complex and emergent behaviors \citep{ca2003}.
Neural Cellular Automata (NCA) extend the classical CA framework by employing a neural network to define the update rule.
This neural network can be trained using gradient descent, allowing NCA to learn complex dynamics from data.
This has led to promising applications in texture synthesis \citep{niklasson2021self-sothtml, meshnca, dynca},
modeling morphogenesis \citep{mordvintsev2020growing},
image generation \citep{vnca, gannca, selfattentionnca}, and discriminative tasks \citep{randazzo2020self-classifying, image-seg-nca}.

NCA\footnote{In this paper, we use NCA to refer to both Neural Cellular \textbf{Automata} and Neural Cellular \textbf{Automaton}.} models operate on a discrete grid of cells, where each cell's state evolves over time according to a local update rule.
The NCA update rule is inspired by Partial Differential Equations (PDEs) that describe reaction-diffusion
systems \citep{turing1990chemical} and can be viewed as an Euler integration of \rev{a} PDE over a discretization of space and time.
A key question surrounding NCA is whether it can truly capture underlying continuous dynamics or simply
overfits the discrete representations used during training.
We investigate this question by studying NCA behavior across different spatio-temporal granularities
and at the limit where space-time discretization approaches continuity. In this work, we focus on NCA models trained for texture synthesis.

\rev{We find that existing NCA models \citep{niklasson2021self-sothtml, dynca}, depending on the target texture, can fail to generalize to spatio-temporal granularities that are different from the training discretization.}
We hypothesize that the overfitting occurs due to NCA's reliance on stochastic updates to break the initial symmetry between the cells.
We propose a simple, yet effective solution to this problem by removing the stochastic updates from the NCA architecture and initializing the cell states with random uniform noise. We call this approach \textit{Noise-NCA}.

Through qualitative and quantitative evaluations, we show that \textit{Noise-NCA} significantly improves the consistency of NCA dynamics across different granularities of space and time discretization.
Furthermore, our improved NCA model unlocks two new test-time interaction capabilities: continuous control over the
speed of pattern formation and the scale of the synthesized textures, \rev{as shown in Figure~\ref{fig:teaser}}.
We demonstrate these functionalities in an online demo available at \url{https://noisenca.github.io}.
Our work reveals \rev{NCA's} ability to learn continuous dynamics, opening new avenues for exploring them
from a dynamical \rev{system's} perspective.

\vspace{-10pt}
\section{Related Works}
\vspace{-1pt}
\subsection{Neural Cellular Automata}
Inspired by Alan Turing's seminal work on Reaction-Diffusion (RD) systems \citep{turing1990chemical}, \cite{mordvintsev2020growing} propose Neural Cellular Automata as a variant of RD systems. Unlike traditional RD systems that require manual design and parameter tuning to create a desired pattern, NCAs can be trained using gradient descent. The authors draw a parallel between NCA and dynamical systems to motivate their design choices. \cite{niklasson2021self-sothtml} further strengthen this relationship by formalizing the NCA training as numerically solving a PDE defined on the cell states. NCA models have been successfully applied in various domains \citep{meshnca, dynca, mordvintsev2021mu-micronca, vnca, selfattentionnca, randazzo2020self-classifying, image-seg-nca}.  
% Following this, researchers have successfully applied NCA in various domains, including texture synthesis \citep{meshnca, dynca, mordvintsev2021mu-micronca}, image generation \citep{vnca, gannca, selfattentionnca}, and discriminative tasks \citep{randazzo2020self-classifying, image-seg-nca}. 
% Despite these advances, most studies employing NCA have merely treated the model as a numerical method for solving the underlying dynamics, with barely any attention paid to whether NCA captures continuous dynamics or overfits discrete representations. 
\rev{Despite these advances, most studies employing NCA have barely paid any attention to whether the model truly captures continuous dynamics or overfits to the training discrete representation.}

\vspace{-3pt}
\subsection{NCA as Partial Differential Equations}
\cite{diffrd} first investigate whether the spatio-temporal discretization used to train their differentiable RD system can effectively approximate a continuous PDE. However, their model lacks a directional spatial gradient and only involves local isotropic information. This sets them apart from common NCA models \citep{mordvintsev2020growing, niklasson2021self-sothtml, meshnca, dynca} where cells perceive their neighborhood in a direction-sensitive manner. 
Moreover, they do not provide a solution for the cases where the model diverges from the expected behavior under the governing PDE. 
\cite{async} explore the Growing NCA model \citep{mordvintsev2020growing} from the perspective of asynchronicity in the update rule and find that a globally synchronous update scheme negatively impacts the temporal continuity of NCA. However, the range of temporal discretizations examined in their work is very limited, leaving the behavior of NCA at the continuous-time limit as an open question. Furthermore, their findings indicate that NCA models, even with asynchronous updates, are prone to overfitting the training discretization. We extend upon previous analyses by pushing NCA to the continuous space-time limit and introduce a simple yet effective solution to improve the spatio-temporal continuity of NCA models. 

\vspace{-4pt}
\section{Method}
\newcommand{\State}{\mathrm{S}}
\newcommand{\Perception}{\mathrm{Z}}
\newcommand{\NState}{\mathrm{S}^{\odot}}

\newcommand{\Kx}{K_\textup{x}}
\newcommand{\Ky}{K_\textup{y}}
\newcommand{\Klap}{K_\textup{lap}}
\newcommand{\Kid}{K_\textup{id}}
\newcommand{\Wone}{\textup{W}_1}
\newcommand{\Wtwo}{\textup{W}_2}
\newcommand{\bone}{\textup{b}_1}
\newcommand{\Dx}{\Delta \textup{x}}
\newcommand{\Dy}{\Delta \textup{y}}
\newcommand{\Dt}{\Delta \textup{t}}

Neural Cellular Automata (NCA) models are a type of Cellular Automata models distinguished by the special characteristic that their update rules are governed by neural networks. In the following sections, we review NCA and introduce the notation that will be used later. In this article, we focus on 2D NCA models; nevertheless, a similar analysis to ours can be performed on other variants of NCA models.

\vspace{-3pt}
\subsection{Definitions}
NCA models operate on a group of cells arranged in a 2D grid. These cells change over discrete time steps according to the NCA update rule. Let $\State(x, y, t) \in \mathbb{R}^C$ denote the $C$ dimensional vector that describes the state of a cell at location $(x, y)$ at time $t$. Note that 
\vspace{-2pt}
$$(x, y) \in \left \{0, 1, \cdots, H - 1 \right \} \mathbf{\times } \left \{0, 1, \cdots, W- 1 \right \}$$
where $H, W$ are the height and width of the grid, respectively. 
% The first 3 dimensions of the $C$ dimensional state represent the RGB color of the cell.
We define $\NState(x, y, t)$ as a set that contains the state of all cells in the neighborhood of a given cell \footnote{The neighborhood can also contain the cell itself.}. The most common neighborhood used in NCA models is the 9-point Moore neighborhood. 

\vspace{-3pt}
\subsection{NCA Update Rule}
As in any Cellular Automata, the NCA update rule is \textit{local} and \textit{space/time invariant}. Locality implies that, at each update step, a cell can only be influenced by a limited number of cells, namely its neighboring cells. Space-Time invariance means that all cells follow the same update rule, which does not change over time.
\cite{meshnca} suggested that the NCA update rule can be decomposed into \textit{perception} and \textit{adaptation} stages. 

% NCA update rule can be written as
% \begin{equation}
% \State(x,y,t+1) = \State(x,y,t) + F_{\theta}(\NState(x,y,t))
% \label{eq:update}
% \end{equation}
% where $F_{\theta}(.)$ denotes the residual update function with parameters $\theta$.
\vspace{-3pt}
\subsubsection{Perception Stage}
\begin{figure}[t!]
    \centering
    \includegraphics[width=\linewidth, trim={0 0pt 0 8pt},clip]{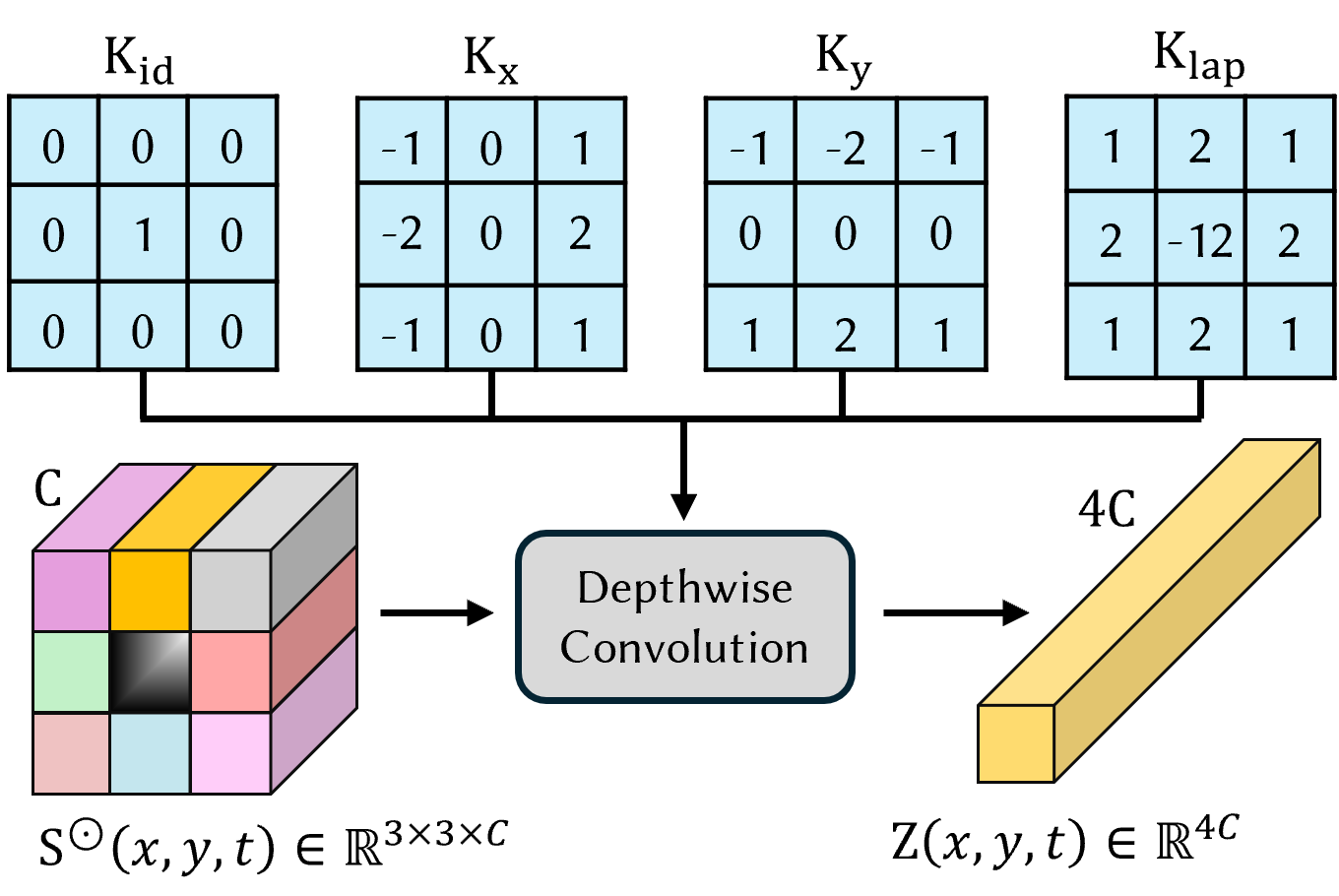}
    \caption{\rev{Overview of the \textit{perception} stage in NCA models for a single cell. Depthwise convolution with the shown filters results in a 4x expansion of the number of channels in the perception vector $\Perception$.}}
    \label{fig:perception}
    \vspace{-9pt}
\end{figure}

In this stage, \rev{as shown in Figure~\ref{fig:perception}}, each cell gathers information from its neighborhood to form its perception vector $\Perception(x,y,t)$ by applying depth-wise convolution on the state of cells in its neighborhood $\NState(x,y,t)$: 
\begin{equation}
\Perception(x,y,t) = \left [ \Kid, \Kx, \Ky, \Klap \right ] \ast \NState(x,y,t)
\label{eq:perception}
\end{equation}
where $\Kid, \Kx, \Ky, \Klap$ denote Identity, Sobel-X, Sobel-Y, and Laplacian filters. \rev{Note that the perception stage is non-parametric, meaning that the convolution filters are frozen during training}. These filters can be modified in test time to interactively control the behavior of cells \citep{niklasson2021self-sothtml, meshnca, dynca}.

\vspace{-3pt}
\subsubsection{Adaptation Stage} 
\begin{figure}[t!]
    \centering
    \includegraphics[width=\linewidth, trim={14pt 0pt 19pt 8pt},clip]{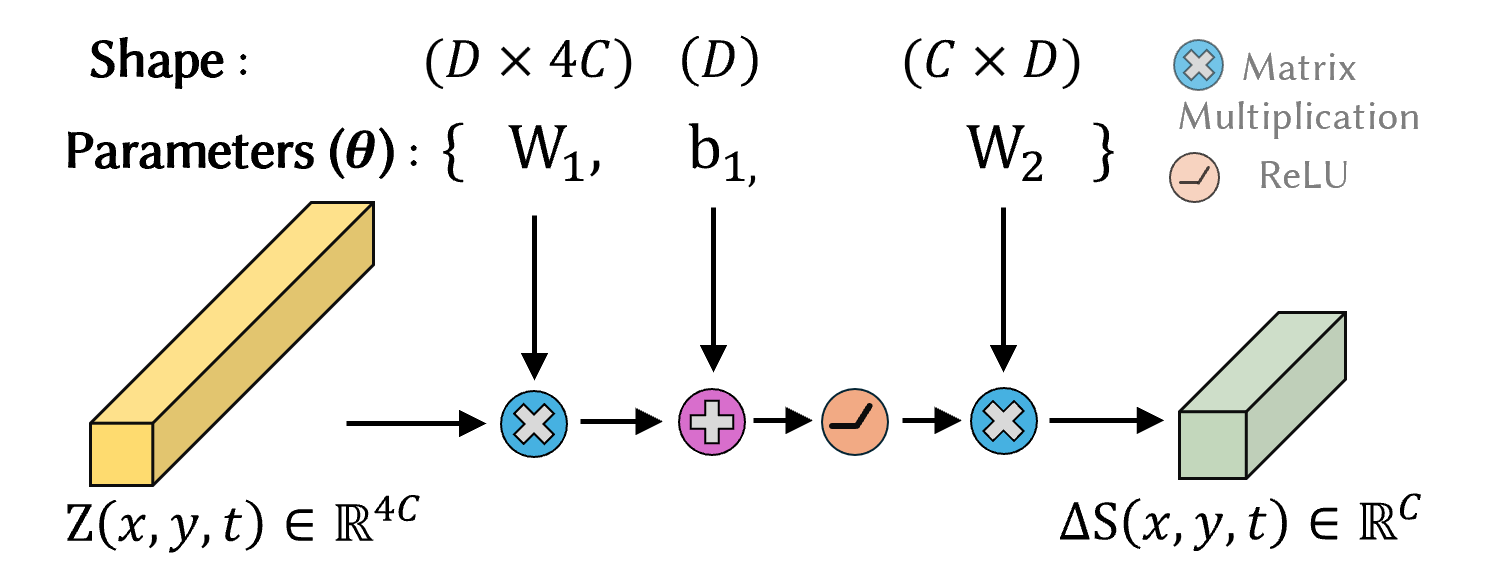}
    \caption{Overview of the \textit{adaptation} stage in NCA models for a single cell. The perception vector goes through a series of operations to compute the residual cell update. $C$ and $D$ denote the number of channels in the cell state and the dimensionality of hidden layer, respectively. }
    \label{fig:adaptation}
    \vspace{-9pt}
\end{figure}

In the adaptation stage, cells can only rely on their perception vector and do not have access to the state of their neighbors. The perception vector passes through a neural network with two layers and a ReLU non-linearity, as shown in Figure~\ref{fig:adaptation}. NCA parameters $(\theta)$ include the weights and bias of the first linear layer $\Wone, \bone$, and the weights of the second linear layer $\Wtwo$. The output of the adaptation stage is the residual update for each cell:
\begin{equation}
\Delta \State(x,y,t) = \Wtwo \left ( \Wone Z(x,y,t) + \bone \right )_{+}
    \label{eq:adaptation}
\end{equation}
where $(.)_+$ denotes the ReLU non-linearity. The final equation describing the cell update can be written as
\begin{equation}
\State(x,y,t+\Dt) = \State(x,y,t) + \Delta \State(x,y,t) \delta(x,y,t) \Dt
\label{eq:update}
\end{equation}
where $\delta(x,y,t) \sim \textup{Bernoulli}(\frac{1}{2})$ is a binary random variable. Multiplying the residual updates by $\delta$ breaks the symmetry between the cells and allows creating new textures over time by introducing randomness into the updates. 
\vspace{-3pt}

\vspace{-2pt}
\subsection{Update Rule As a PDE}
The NCA update rule in Equation~\ref{eq:update} can be viewed as an Euler integration over a \textbf{spatio-temporal discretization of a continuous PDE} \footnote{Technically a Stochastic Differential Equation (SDE) due to presence of $\delta(x,y,t)$ in the update rule.} of the following form: 
\begin{equation}
        \frac{\partial \mathrm{S}}{\partial t} = f_{\mathrm{\theta}} (\State, \nabla_{\mathrm{x}}\State, \nabla_{\mathrm{y}} \State, \nabla^2 \State )
        \label{eq:pde}
\end{equation}
In the \textit{Perception} stage, the spatial derivatives including $\nabla_{\mathrm{x}}, \nabla_{\mathrm{y}}$, and $\nabla^2$ are estimated using Sobel filters $\Kx, \Ky$ and 9-point discrete Laplacian kernel $\Klap$, respectively. In the \textit{Adaptation} stage, the cell states are updated \rev{by applying} Euler integration with timestep $\Dt$.  Note that the Sobel and Laplacian filters used in the \textit{Perception} stage need to be divided by the cell size $\Dx, \Dy$ to get consistent estimations of the spatial derivatives across different resolutions. According to the finite difference method, a consistent approximation of spatial gradients using Sobel filters can be achieved using: 
\vspace{-11pt}
$$ \nabla_{\mathrm{x}}, \nabla_{\mathrm{y}} \approx  \frac{\Kx}{\Dx}, \frac{\Ky}{\Dy}$$

For the Laplacian operator, we first need to decompose it into second-order derivatives with respect to $x, y$, which can be written as $\nabla^2 = \frac{\partial^2 }{\partial x^2} + \frac{\partial^2 }{\partial y^2}$. Similarly, the Laplacian convolution kernel can be decomposed to two terms:
\vspace{-4pt}
$$
\begin{aligned}
\;\;\;\;\;\;\; \Klap \;\;\;\;\;\;\; = \;\;\;\;\;\;\;\;\;\; \Klap^\mathrm{x} \;\;\;\;\;\;\;\;\;\;+ \;\;\;\;\;\;\;\; \;\; \Klap^\mathrm{y} \;\;\;\;\;\;\;\;\;
\\
\begin{bmatrix}
1 & 2 & 1\\ 
2 & -12 & 2\\ 
1 & 2 & 1
\end{bmatrix} = 
\begin{bmatrix}
0.5 & 0 & 0.5\\ 
2 & -6 & 2\\ 
0.5 & 0 & 0.5
\end{bmatrix} + 
\begin{bmatrix}
0.5 & 2 & 0.5\\ 
0 & -6 & 0\\ 
0.5 & 2 & 0.5
\end{bmatrix}
\end{aligned}
$$
Therefore a discretization-consistent approximation of the Laplacian can be written as:
\vspace{-5pt}
$$\nabla^2 \approx  \frac{\Klap^\mathrm{x}}{\Dx^2} + \frac{\Klap^\mathrm{y}}{\Dy^2}$$

Combining the different pieces to make the NCA perception stage consistent across different spatial granularities, the convolution filters need to be modified as follows: 

\vspace{-10pt}
\begin{equation}
\left [ \Kid, \Kx, \Ky, \Klap \right ] \rightarrow \left [ \Kid, \frac{\Kx}{\Dx}, \frac{\Ky}{\Dy},\frac{\Klap^\mathrm{x}}{\Dx^2} + \frac{\Klap^\mathrm{y}}{\Dy^2} \right ]
\label{eq:pde_perception}
\end{equation}
\vspace{-6pt}

The cell size $(\Dx, \Dy)$ and timestep $\Dt$ are all set to $1.0$ during training, but can be changed in test time to evaluate the NCA model across different spatial and temporal sampling granularities. Ideally, we want the learned NCA update rule to correspond to the PDE in Equation~\ref{eq:pde} and to be able to generalize to different $\Dt$ and $(\Dx, \Dy)$ in test time.

\vspace{-3pt}
\subsection{Baseline NCA models}
We analyze and experiment with two NCA baselines. The first baseline is the model proposed by \cite{niklasson2021self-sothtml}. Figures~\ref{fig:perception} and \ref{fig:adaptation} accurately depict its architecture, which we refer to as \textit{Vanilla-NCA}.
The second baseline is adapted from the DyNCA model by \cite{dynca}. We refer to it as \textit{PE-NCA} where PE is short for \textbf{p}ositional-\textbf{e}ncoding. \textit{PE-NCA} appends the normalized cell coordinates $(\frac{x}{H}, \frac{y}{W})$ to the perception vector $Z(x,y,t)$, which allows the cells to be aware of their position and thus making $\Wone$ a $(D\times 4C + 2)$ matrix. Note that \textit{Vanilla-NCA} uses circular padding when applying convolution kernels in the perception stage, while \textit{PE-NCA} uses replicate padding to make the boundary condition consistent with positional encoding. In both of these baselines, the initial state\footnote{Also called seed}, is set to zero $\State(x,y,t=0)=0.0$. The \textit{Vanilla-NCA} model relies only on the stochastic update mask $\delta(x,y,t)$ to break the symmetry between cells in the initial state $t=0$, while the positional encoding in \textit{PE-NCA} acts as another source to differentiate between cells. 

\vspace{-4pt}
\subsection{Proposed Model: NoiseNCA}
We propose a simple yet effective change to the baseline NCAs that will prevent the model from overfitting the training discretization. Our proposed NCA model, namely \textit{Noise-NCA} is similar to the \textit{Vanilla-NCA} with the main difference being that we use random uniform noise as the initial state, i.e., seed
\vspace{-6pt}
\begin{equation}
    \State(x,y,t=0) \sim \mathcal{U}[-\epsilon, \epsilon]
\end{equation}
where $\epsilon$ represents the strength of the noise. By setting the initial condition to random noise, \textit{Noise-NCA} perfectly breaks the symmetry between cells and eliminates the need for stochastic update in the update rule. Therefore, \textit{Noise-NCA}'s update rule is deterministic and can be written as: 
\begin{equation}
\State(x,y,t+\Dt) = \State(x,y,t) + \Delta \State(x,y,t) \Dt.
\label{eq:noise_update}
\end{equation}

In the experiments section we evaluate all three NCA variants to see whether the learned update rule of each model corresponds to a continuous dynamic described by the PDE in Equation~\ref{eq:pde} or merely overfits to the training discretization. Our approach is to analyze the behavior of NCA models at the continuous limit where the temporal $\Dt \to 0^+$ and spatial $\Dx, \Dy \to 0$ discretizations become more and more fine-grained. 
Our experiments demonstrate the advantages of \textit{Noise-NCA} over the baselines through qualitative and quantitative evaluations. 

\vspace{-4pt}
\subsection{Training Scheme}
\label{sec:train}
We use the same training strategy to train all three NCA variants: \textit{Vanilla-NCA}, \textit{PE-NCA}, and our \textit{Noise-NCA}.
% We focus on NCA models that are trained to create textures and patterns. 
To train the model, we start from an initial state $\State(x,y,t=0)$ and iteratively apply the NCA update rule to obtain cell states $\State(x,y,t)$ at time $t$. The RGB image is then extracted by taking the first 3 channels of $\State(x,y,t)$ and compared to the target texture using a loss function. We use the appearance loss function proposed by \cite{dynca} which measures the similarity of two texture images using a pre-trained VGG16 network \citep{vgg}. The gradients from the loss function are backpropagated through time and used to optimize the NCA parameters $(\theta)$. All models are trained at resolution of $H,W=(128,128)$. 
We use the pool technique proposed by \cite{mordvintsev2020growing} to improve the long-term stability of the update rule.

% \ep{More training details if space permits} \yx{can put into supp. Check the name of the supp file maybe Alife doesn't call it supp}
% Notice that both of ours baselines use $\Delta t = 1.0$ in the training and testing, so the model is only trained and evaluated on integer timesteps $t \in \left \{0, 1, 2, \cdots \right \}$.

% Whether the approximation of the time derivative and spatial gradients lead to actual continuous dynamic remains a question. In the following sections, we study the NCA in continuous-time limit and continuous-space limit to answer it.

% For an NCA to be a well-behaved PDE, it should adapt to different dt and spatial scales during test-time. 

% in the next sections we study the NCA in the limit

\vspace{-3pt}
\section{Experiments}
We train the three NCA variants - \textit{Vanilla-NCA}, \textit{PE-NCA}, \textit{Noise-NCA} - on 45 different textures taken from \cite{dynca} using the training scheme described previously.
We conduct qualitative and quantitative experiments to evaluate the trained NCA models at continuous-time limit $\Dt \to 0^+$ and continuous-space limit $\Dx,\Dy \to 0$. In the paper, the qualitative results are demonstrated for two textures: "bubbly\_0101", and "grid\_0135". The complete qualitative results, which compare different NCA variants on all 45 textures, are available at \url{https://noisenca.github.io/supplementary/}.
For a quantitative evaluation, we test the models on all of the 45 textures and use the appearance loss proposed by \cite{dynca} as a metric to quantify the quality of the NCA output for different values of $\Dt$ and $\Dx, \Dy$. 

\vspace{-3pt}
\subsection{Continuous-Time Limit}
\begin{figure}[t!]
    \centering
    \includegraphics[width=\linewidth, trim={10pt 0pt 0pt 4pt},clip]{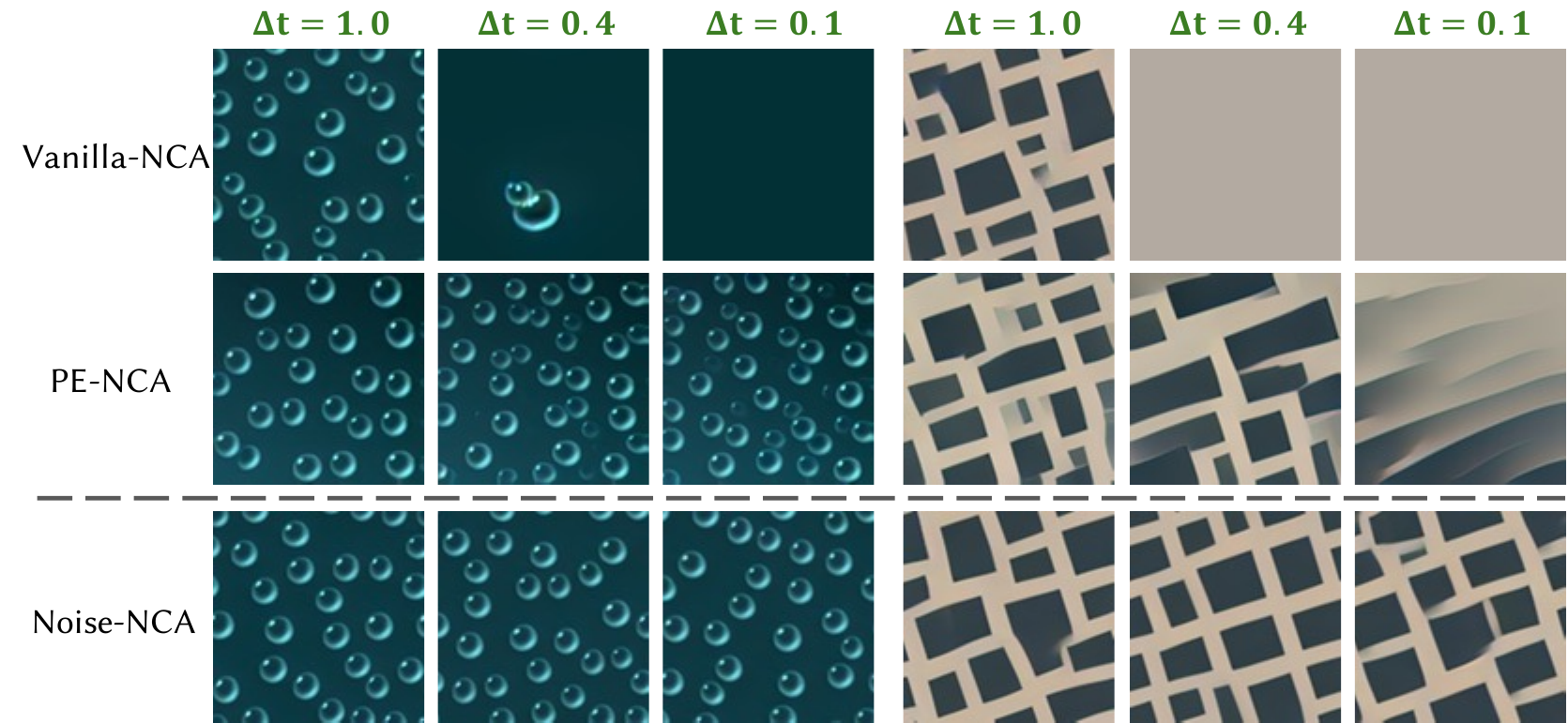}
    \caption{Output of NCA models for different values of $\Dt$ used in the Euler integration, on two different textures.
    As $\Dt$ decreases, the patterns created by \textit{Vanilla-NCA} and \textit{PE-NCA} start to deviate from the
    correct pattern, while \textit{Noise-NCA} is able to maintain the correct pattern for  all $\Dt$ values.}
    \label{fig:dt_qualitative}
    \vspace{-9pt}
\end{figure}

Given a $\Dt \leq 1.0$ and a simulation time $T$, we start from the initial state $\State(x,y, t=0)$
and use Euler integration to find $\State(x,y,t=T)$ by applying the NCA update rule for $\left \lceil \frac{T}{\Dt} \right \rceil$ steps. We set $T=300$ in our experiments.
Figure~\ref{fig:dt_qualitative} shows the output of the three NCA variants on two different textures for $\Dt \in \{1.0, 0.4, 0.1\}$.
As shown in the figure, when $\Dt=1.0$ all models successfully create the correct patterns. However, as we decrease $\Dt$ the \textit{Vanilla-NCA} and \textit{PE-NCA} fail to synthesize the correct texture. On the contrary, \textit{Noise-NCA} is able to successfully generate the correct pattern across all values of $\Dt$.

\begin{figure}[t!]
    \centering
    \includegraphics[width=0.95\linewidth, trim={9pt 5pt 15pt 12pt},clip]{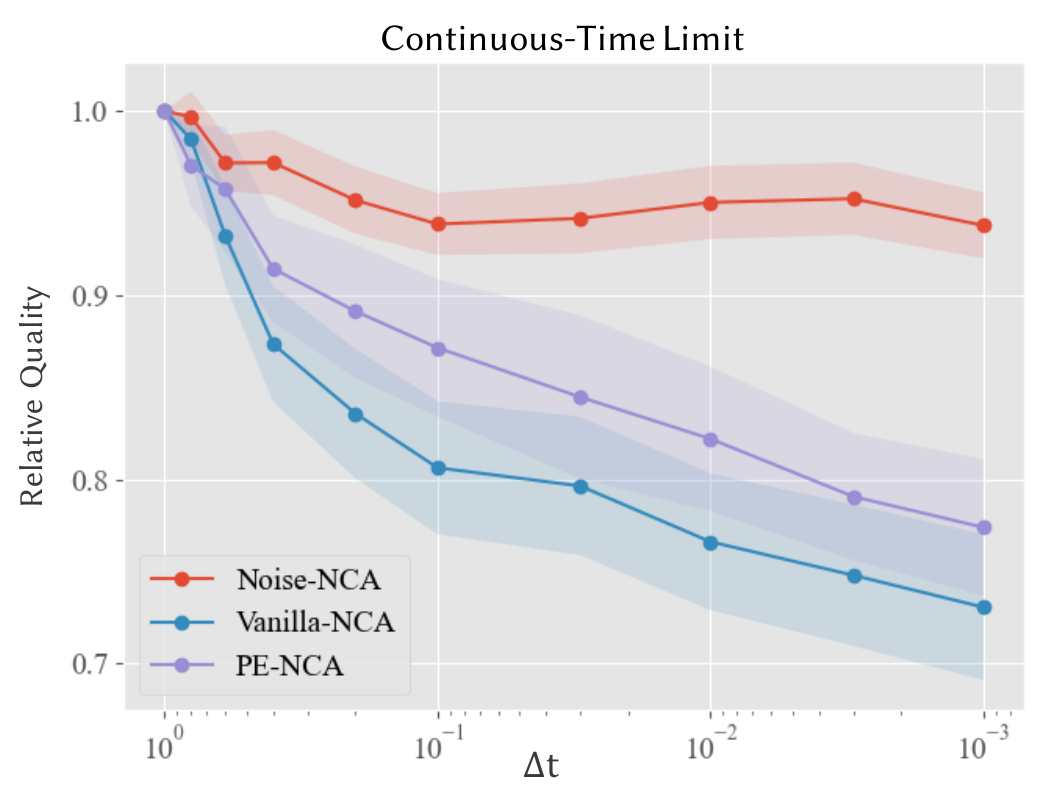}
    % \vspace{-3pt}
    \caption{For each NCA variant, we evaluate the relative output quality for different values of $\Dt$ used for Euler integration
    and show the results averaged over 45 different textures. \textit{Noise-NCA} shows significantly better
    generalization to different timesteps compared to the baseline NCAs, even for very small values of $\Dt$.}
    \label{fig:dt_quantitative}
    \vspace{-10pt}
\end{figure}

For a more rigorous comparison, we perform a quantitative evaluation on 45 different textures \rev{for ten different} $\Dt$ values ranging from $[10^{-3}, 10^0]$.
Let $\State^{\Dt}_T$ denote the output of NCA at time $T$ given $\Dt$ as the integration timestep.
For each value of $\Dt$, we evaluate 
\vspace{-5pt}
$$\frac{\mathcal{L}(\State^{1.0}_T)}{\mathcal{L}(\State^{\Dt}_T)}$$ where $\mathcal L$ is our appearance loss function.
This metric allows us to quantify, for each NCA variant, the relative output quality
when the inference timestep $\Dt$ is different from the training
timestep $\Dt=1.0$.
The results in Figure~\ref{fig:dt_quantitative} demonstrate
that while the baseline NCAs fail to generalize to timesteps smaller than
$\Dt \leq 1.0$, \textit{Noise-NCA} is able to maintain a consistent output quality even for very small values of $\Dt$.
This suggests that \textit{Noise-NCA} is able to learn a continuous dynamic described by the PDE in Equation~\ref{eq:pde}.

\vspace{-3pt}
\subsection{Continuous-Space Limit}
We evaluate the NCA models at the continuous-space limit by increasing the spatial resolution of the grid.
In our experiments, we set $\Dx = \Dy$ for simplicity and use a square grid with size $H = W = \left \lceil \frac{128}{\Dx} \right \rceil$,
with $128\times128$ being the resolution used during training. To visualize and compare the results,
we resize the NCA output from $H \times W$ to $128 \times 128$.
Notice that, as shown in Equation~\ref{eq:pde_perception}, for $\Dx \leq 1.0$ the output of the Sobel and Laplacian filters are magnified by $\frac{1}{\Dx}$ and
$\frac{1}{\Dx^2}$, respectively. Therefore, to avoid Euler integration from overshooting, we set $\Dt = min(1.0, \Dx^2)$.

\begin{figure}[t!]
    \centering
    \includegraphics[width=\linewidth, trim={10pt 0pt 0pt 4pt},clip]{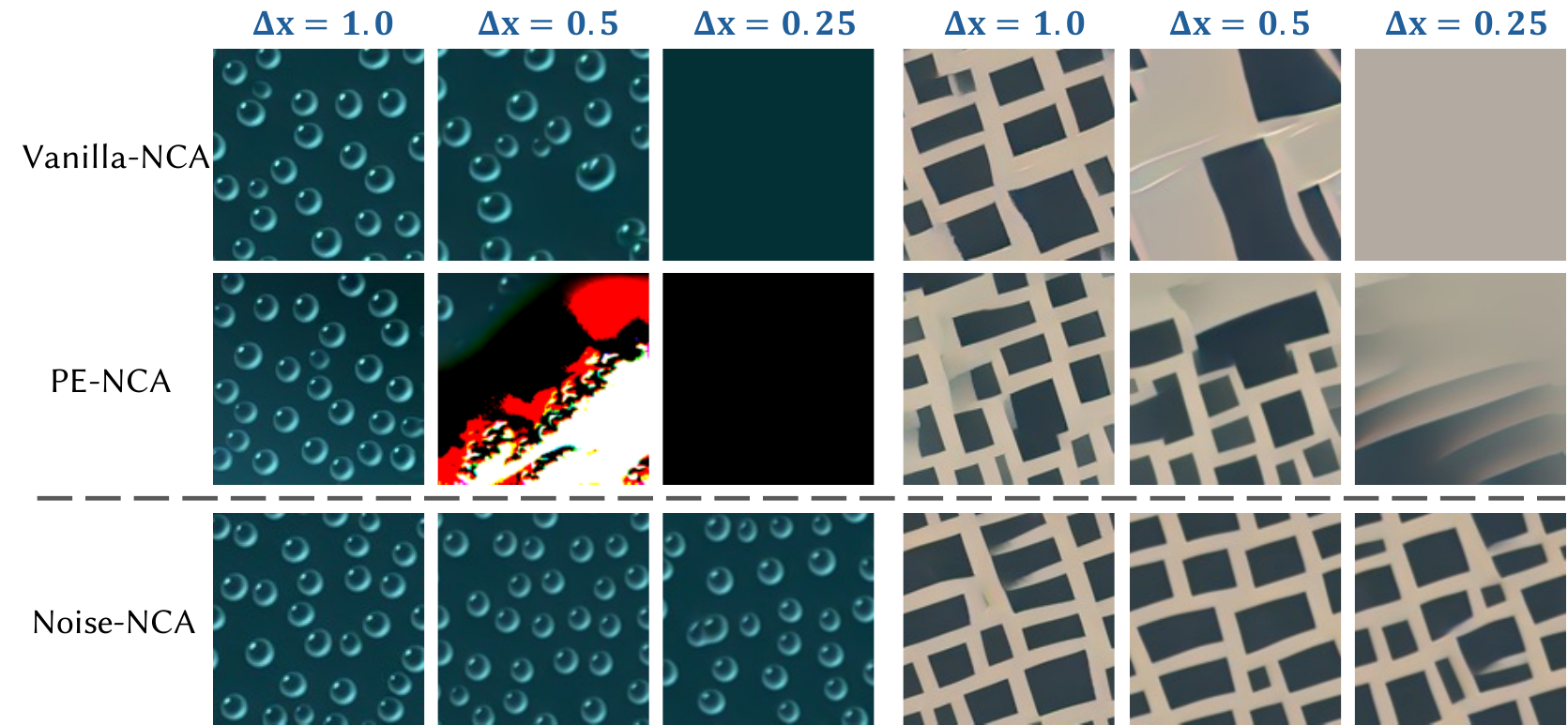}
    % \vspace{-3pt}
    \caption{Output of NCA models for different resolutions with varying spatial discretization granularity $\Dx$.
    As $\Dx$ decreases, the patterns created by \textit{Vanilla-NCA} and \textit{PE-NCA} start to deviate from the
    correct pattern, while \textit{Noise-NCA} is able to maintain a consistent output.}
    \label{fig:dx_qualitative}
    \vspace{-4pt}
\end{figure}

\begin{figure}[t!]
    \centering
    \includegraphics[width=0.95\linewidth, trim={9pt 5pt 11pt 10pt},clip]{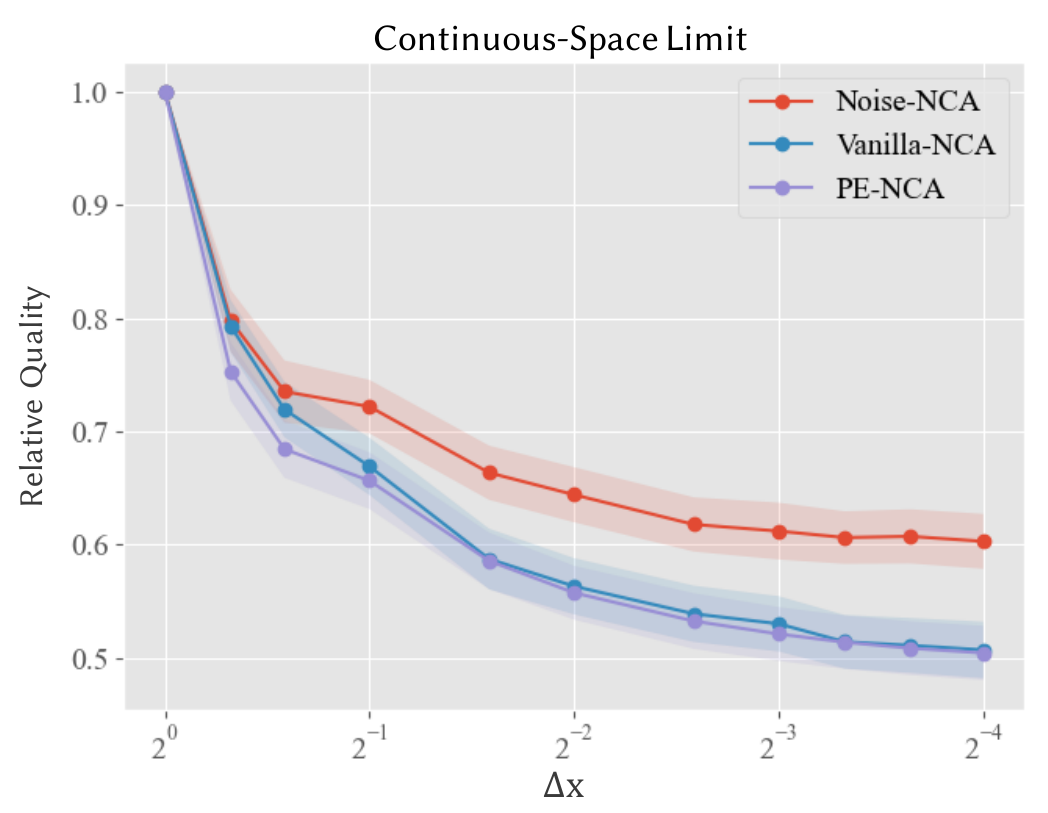}
    \caption{We simulate the NCA model on different spatial resolutions and evaluate the relative output quality
    for different values of $\Dx$ and show the results averaged over 45 textures.
    \textit{Noise-NCA} shows improved generalization to different spatial resolutions compared to the baseline NCAs.}
    \vspace{-4pt}
    \label{fig:dx_quantitative}
\end{figure}

Figure~\ref{fig:dx_qualitative} shows the output of all three NCA variants on two different textures for $\Dx \in \{1.0, 0.5, 0.25\}$.
The qualitative results in the figure show that \textit{Noise-NCA} is able to generate the correct pattern across all
values of $\Dx$, while the baseline NCAs fail to generalize to different spatial resolutions.
For a quantitative evaluation of the effect of $\Dx$ on the output quality,
we use eleven different  $\Dx$ values ranging from $[2^{-4}, 2^0]$, which results in grids with resolutions varying
from $2048\times2048$ to $128\times128$ and test the models on all 45 textures.
For a given value of $\Dx$, let $\State^{\Dx}_T$ denote the output of NCA at time $T$ given cell size $\Dx$ and let $\downarrow_{128}$ denote the bilinear interpolation operator for resizing to resolution $128 \times 128$.
As a measure of relative output quality, we evaluate the ratio of appearance losses \vspace{-0pt} $$\frac{\mathcal{L}(\State^{1.0}_T)}{\mathcal{L}(\downarrow_{128} \State^{\Dx}_T)}$$ for each NCA variant and show the results in Figure~\ref{fig:dx_quantitative}.
The quantitative results show that \textit{Noise-NCA} is better able to generalize to different spatial resolutions and maintains a higher quality compared to the baseline NCAs. 
% \ep{Explain why there is a gap still for the spatial results.}

Our experiments demonstrate that  \textit{Vanilla-NCA} and \textit{PE-NCA} overfit the spatial and temporal resolution used during training.
This indicates that the dynamics learned by the baseline NCAs do not correspond to a
continuous PDE, as they exhibit inconsistent behavior during training $(\Dt, \Dx, \Dy = 1.0)$ and inference $(\Dt, \Dx, \Dy \to 0)$.
On the other hand, the dynamics learned by \textit{Noise-NCA} behave much more consistently across different spatial and temporal resolutions.
We also find that \textit{Noise-NCA} achieves similar results if we modify the update rule at inference time by using alternative differential equation solvers such as the Runge-Kutta methods instead of the Euler method. These results show that our modifications to the NCA architecture improve the continuity of the dynamics learned by our \textit{ Noise-NCA} model and allow it to find an update rule that corresponds to an actual PDE.

\section{Insights into the Overfitting Problem}
In this section, we further investigate NCA and provide insights into the source of the overfitting.
We also study the behavior of NCAs from a fixed point perspective to understand the output of overfitting NCA models at the limit $\Dt \to 0$.

\subsection{Source of Overfitting}
Our experiments in the previous section demonstrate that \textit{Vanilla-NCA} \rev{overfits} the timestep used during training, namely $\Dt = 1.0$.
We analyze this phenomenon and provide intuition that motivates and supports the design of \textit{Noise-NCA}. 
% \yx{motivate here seems a bit too late. how about support/make it sound}
We introduce two different scheduling policies $\Dt_{\mathrm{A}}(t), \Dt_{\mathrm{B}}(t)$ that will determine the timestep value $\Dt$ as a function of time $t$.
$$
\Dt_{\mathrm{A}}(t) = \left\{\begin{matrix}
1.0 & t \leq T_{crit}\\
0.1 & t > T_{crit}
\end{matrix}\right.
\;\;\;
\Dt_{\mathrm{B}}(t) = \left\{\begin{matrix}
0.1 & t \leq T_{crit}\\
1.0 & t > T_{crit}
\end{matrix}\right.
$$

\begin{figure}[htbp]
    \centering
    \includegraphics[width=\linewidth, trim={6pt 0pt 0pt 4pt},clip]{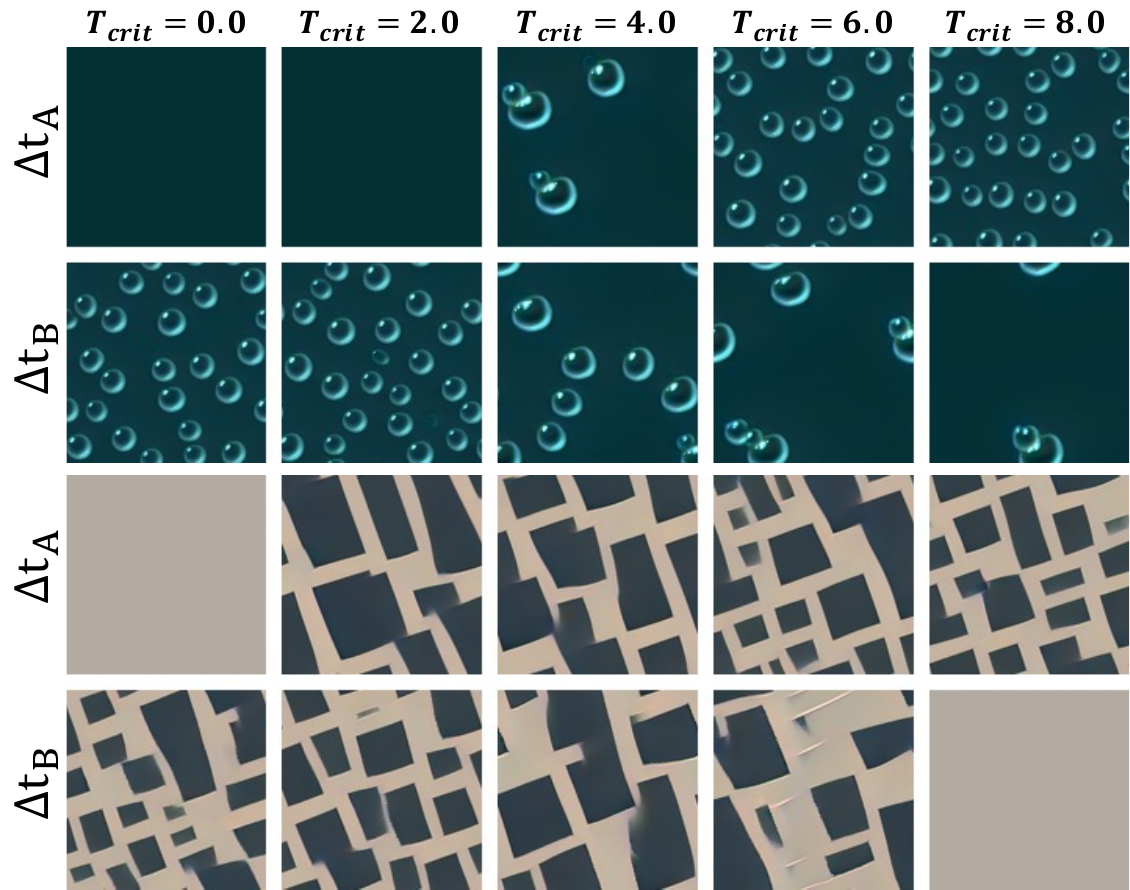}
    \caption{The output of \textit{Vanilla-NCA} at $t=300$ for different values of $T_{crit}$ using two different schedulers $\Dt_{\mathrm{A}}(t), \Dt_{\mathrm{B}}(t)$.}
    \label{fig:dt_scheduler}
\end{figure}

$T_{crit}$ in the above equations indicates the critical time at which the behavior of the scheduler will change. 
Given a scheduler, $A$ or $B$, we use Euler integration to evaluate the output of \textit{Vanilla-NCA} at $t=300$ for different values of $T_{crit}$. Figure~\ref{fig:dt_scheduler}
shows the resulting output of \textit{Vanilla-NCA} for five different values of $T_{crit}$.
Notice that when $T_{crit} = 0.0$, \textit{Vanilla-NCA} can successfully create the correct patterns using scheduler $\mathrm{B}$ as $\Dt_{\mathrm{B}}(t)=1.0$, and fails to create the patterns using scheduler $\mathrm{A}$ since $\Dt_{\mathrm{A}}(t)=0.1$. However, as $T_{crit}$ increases, the output quality for scheduler $A$ improves, while the opposite happens for scheduler $B$. Note that the time required for the patterns to be fully formed is around $t = 100$, and thus the values of $T_{crit}$ used in our experiment are very small compared to pattern formation time.
The results in Figure~\ref{fig:dt_scheduler} suggest that \textit{Vanilla-NCA} relies on the training timestep $\Dt=1.0$ when $t$ is small and the cell states are close to the seed state.
We hypothesize that this overfitting occurs due to NCA's reliance on stochastic updates to break the initial symmetry of the seed state. By replacing the zero seed state with uniform noise, \textit{Noise-NCA} does not rely on stochastic updates to break the initial symmetry between cells and shows significantly improved generalization to unseen timesteps $\Dt$, as shown in Figures~\ref{fig:dt_qualitative}, and \ref{fig:dt_quantitative}.

\subsection{Fixed Point Analysis}
\begin{figure}[t!]
    \vspace{-4pt}
    \centering
    \includegraphics[width=\linewidth, trim={6pt 6pt 0pt 0pt},clip]{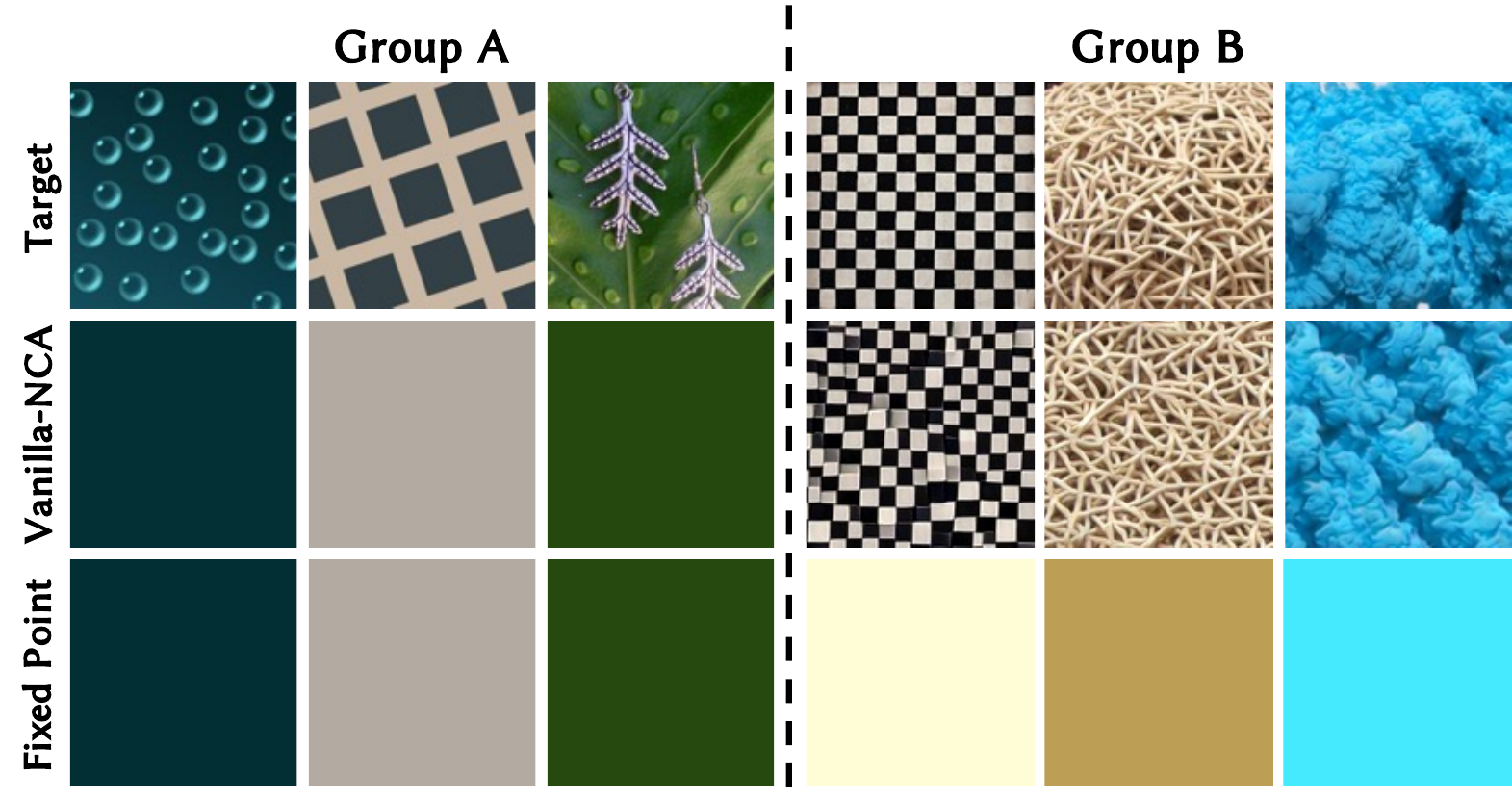}
    \caption{
        First row: Target texture, Second row: Output of \textit{Vanilla-NCA} at $t=300$ using $\Dt=0.001$.
        Third row: Fixed point state of the model obtained by numerically solving the optimization problem in Equation~\ref{eq:fixed_point_opt}.}
    \label{fig:fixed_point}
\end{figure}

We find that at the continuous-time limit $\Dt \to 0^+$, the behavior of the \textit{Vanilla-NCA} model, either \rev{overfitting or generalizing}, largely depends on the target texture. As shown in the second row of Figure~\ref{fig:fixed_point}, when $\Dt$ is small, for textures in Group B, the model is able to create the correct pattern, while for textures in Group A, it converges to a uniform solution that resembles the background color of the target pattern. \rev{We speculate that this phenomenon is related to the presence of solitons in the NCA output as the textures in Group A exhibit soliton like patterns and have a background/foreground separation.} 
% Figure~\ref{fig:fixed_point} shows the output of \textit{Vanilla-NCA} on six different target textures at time $t = 300.0$ using $\Dt=0.001$. 
% We find that at the continuous-time limit $\Dt \to 0$, \textit{Vanilla-NCA} behaves differently depending on the target texture.
In this section, we provide a fixed point analysis of the \textit{Vanilla-NCA} model to better understand the overfitting problem and the nature of the uniform solution that \textit{Vanilla-NCA} converges to.
In this uniform solution, all cells converge to the same state and become stable.
We refer to this state as a fixed point state $\State_{fp} \in \mathbb{R}^C$ of the NCA.
In the CA literature, this state $\State_{fp}$ is often referred to as a \textit{Quiescent State} \citep{sayama2015introduction}.
The property of the quiescent state is that if all neighbors of a cell, including the cell itself, are in the quiescent state,
then the cell will remain in the quiescent state in the next iteration
\footnote{For example in the Game of Life CA, dead states are quiescent states since a dead cell remains dead if all of its neighbors are dead.}.
In the context of NCA, the quiescent state is a local fixed point of the NCA model.

\newcommand{\Zx}{\Perception_\textup{x}}
\newcommand{\Zy}{\Perception_\textup{y}}
\newcommand{\Zlap}{\Perception_\textup{lap}}
\newcommand{\Zid}{\Perception_\textup{id}}

\newcommand{\Wx}{\Wone^\textup{x}}
\newcommand{\Wy}{\Wone^\textup{y}}
\newcommand{\Wlap}{\Wone^\textup{lap}}
\newcommand{\Wid}{\Wone^\textup{id}}

If $\State_{fp}$ is a quiescent state of the NCA model, then the value of $\Delta \State$ in Equation~\ref{eq:adaptation}
should be zero.
Let $\Perception_{fp} = \left [ \Zid, \Zx, \Zy, \Zlap \right ]$ be the perception vector of a cell in the quiescent state.
\rev{Since all neighboring cells are also in the quiescent state, we can simplify the equation by using the fact that
applying the Sobel and Laplacian filters to this neighborhood will yield zero output.} 
Therefore, $\Zx=\Zy=\Zlap = 0$ and $\Zid = \State_{fp}$. Let $\left [ \Wid, \Wx, \Wy, \Wlap \right ]$ be a partitioning
of the weight vector $\Wone$ where each partition is a $D \times C$ matrix that operates on its corresponding part of the perception vector.
To find a fixed point state of an NCA model we propose to solve the following optimization
problem which minimizes the $L_1$ norm of the cell update vector $\Delta \State$:
\begin{equation}
    \State_{fp} = \argmin_{\State \in \mathbb{R}^C} \left \| \Wtwo \left ( \Wid \State+ \bone \right )_{+} \right \|_1
    \label{eq:fixed_point_opt}
\end{equation}
\rev{Note that the optimization objective in Equation~\ref{eq:fixed_point_opt} is convex, and hence all local minimas are connected and every local minima is also a global minima.} To find \rev{a} fixed point state of an NCA, we numerically solve this optimization problem using gradient descent.  \rev{We find that, regardless of the starting point, the optimization converges nearly to the same solution. This suggests that the fixed point states, i.e. the global minimas, are tightly packed.
}
To visualize the obtained fixed point, we set the state of all cells to be equal to this fixed point state and show the results in the third row of Figure~\ref{fig:fixed_point} .
These results show that for textures in Group A, applying the NCA update rule with small $\Dt$ (illustrated in the second row of Figure 9) converges to the same state as the fixed point solution given by Equation~\ref{eq:fixed_point_opt}.
However, for textures in Group B, the model synthesizes the correct pattern using small $\Dt$ values, showing that the NCA output does not converge to the fixed point solution.
We hypothesize that this behavioral difference between Groups A and B is due to the stability of the fixed point state.
In practice, the value of $\Delta \State_{fp}$ is very close to but not exactly zero; therefore, the NCA model can escape from an unstable or saddle fixed point.
We find that if we initialize the cell state with $\State_{fp}$ and apply the NCA update rule for a large number of steps,
the NCA will escape the fixed point state and converge to the correct pattern for textures in Group B, 
while it will remain in the fixed point state for textures in Group A. This shows that the fixed point state is stable
for textures in Group A and is a saddle point for textures in Group B. \rev{We speculate that for textures in Group A, when $\Dt$ is small, the stochastic update does not create enough variation between cells, which causes the NCA to converge to a uniform state.}

%https://math.libretexts.org/Bookshelves/Scientific_Computing_Simulations_and_Modeling/Book%3A_Introduction_to_the_Modeling_and_Analysis_of_Complex_Systems_(Sayama)/11%3A_Cellular_Automata_I__Modeling/11.01%3A_De%EF%AC%81nition_of_Cellular_Automata

\section{Scale of the Patterns}

\begin{figure*}
    \centering
    \includegraphics[width=\textwidth]{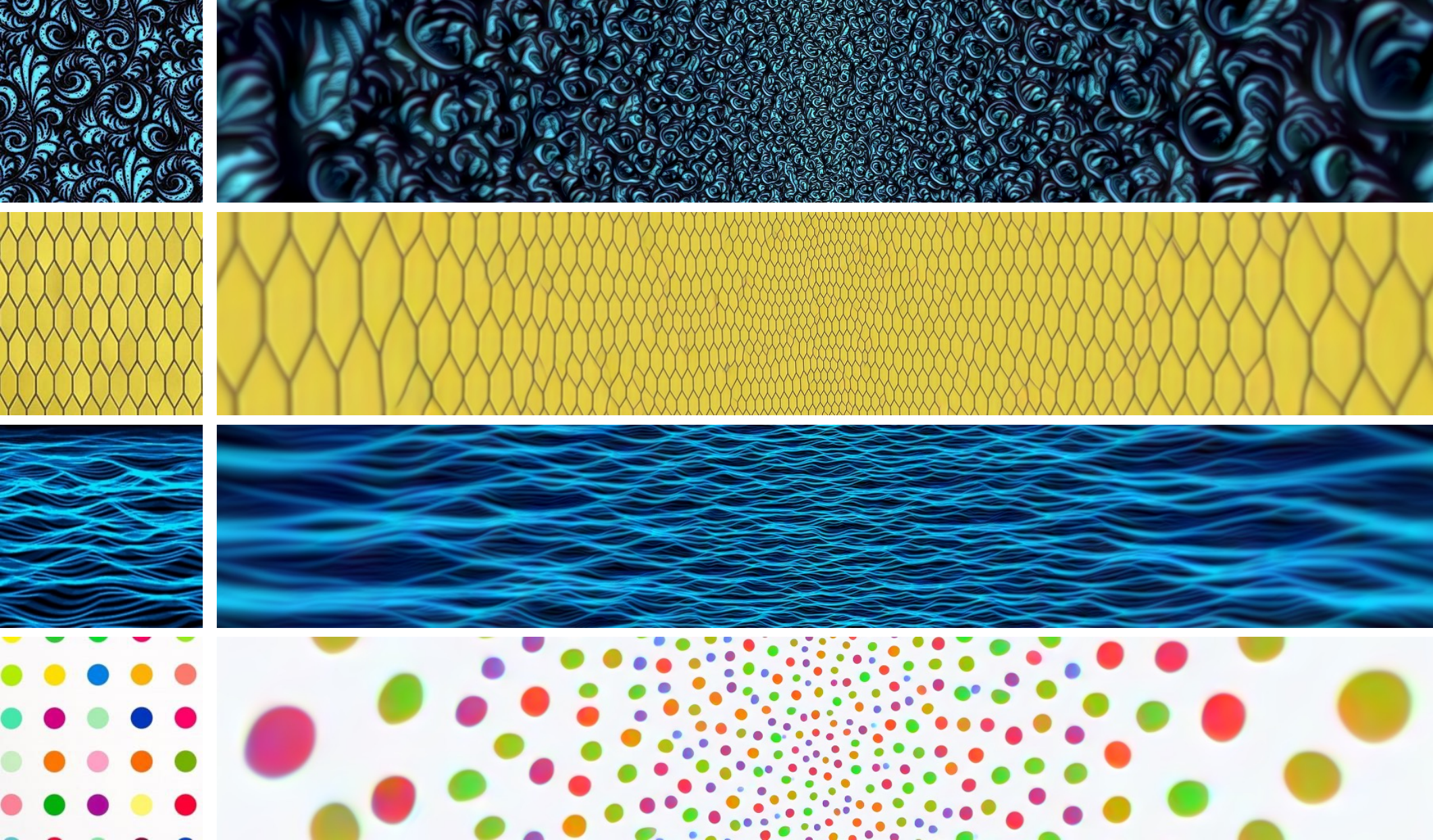}
    \caption{\rev{By varying the cell size $\Dx, \Dy$ as a function of the cell coordinate $x, y$, we can create multiscale patterns.
    The target texture is shown on the left and the synthesized images demonstrating multiscale patterns are shown on the right.}}
    \label{fig:multiscale}
    \vspace{-12pt}
\end{figure*}

\begin{figure}[t!]
    % \centering
    \includegraphics[width=\linewidth, trim={4pt 0pt 0pt 4pt},clip]{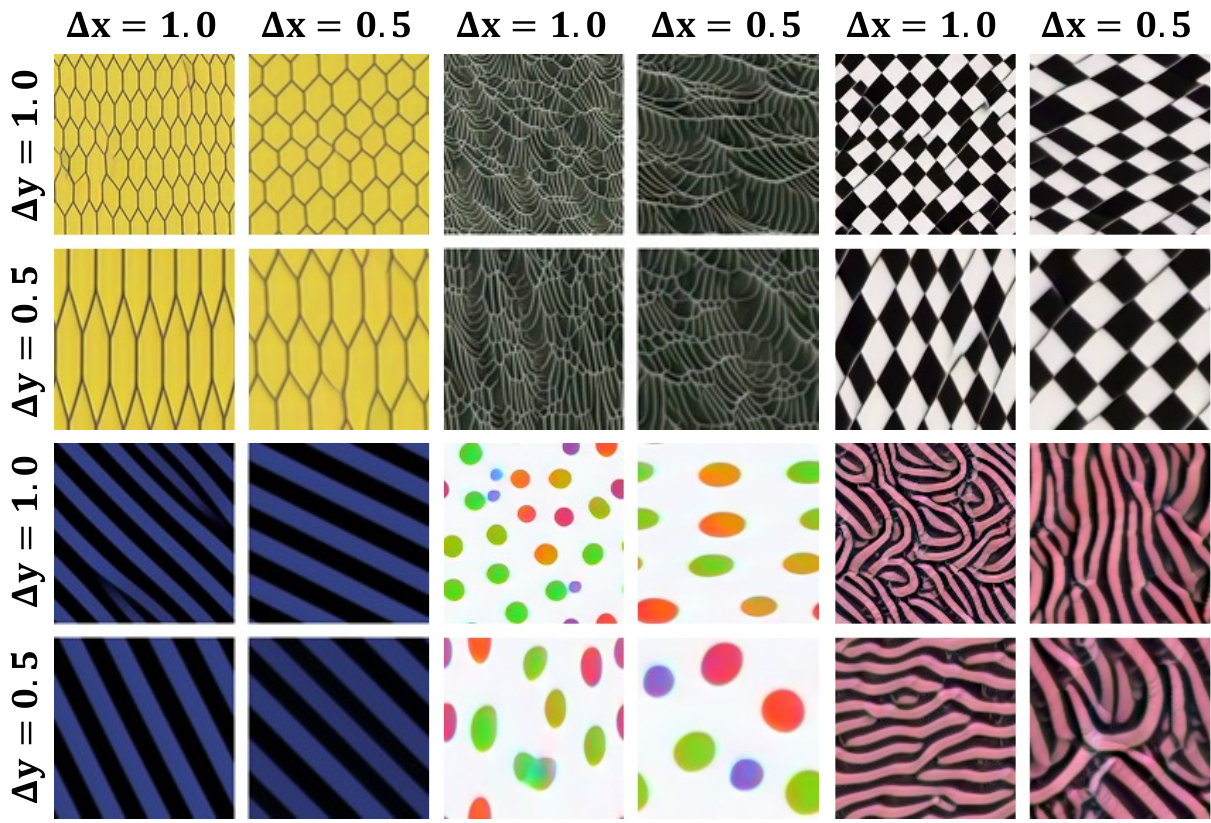}
    \caption{By using different values for $\Dx, \Dy$, \textit{Noise-NCA} allows for anisotropic scaling.
    Decreasing $\Dx, \Dy$ stretches the patterns horizontally and vertically, respectively.
    }
    \label{fig:anisotropic}
    \vspace{-13pt}
\end{figure}

\begin{figure}[t!]
    \centering
    \includegraphics[width=\linewidth, trim={10pt 0pt 0pt 18pt},clip]{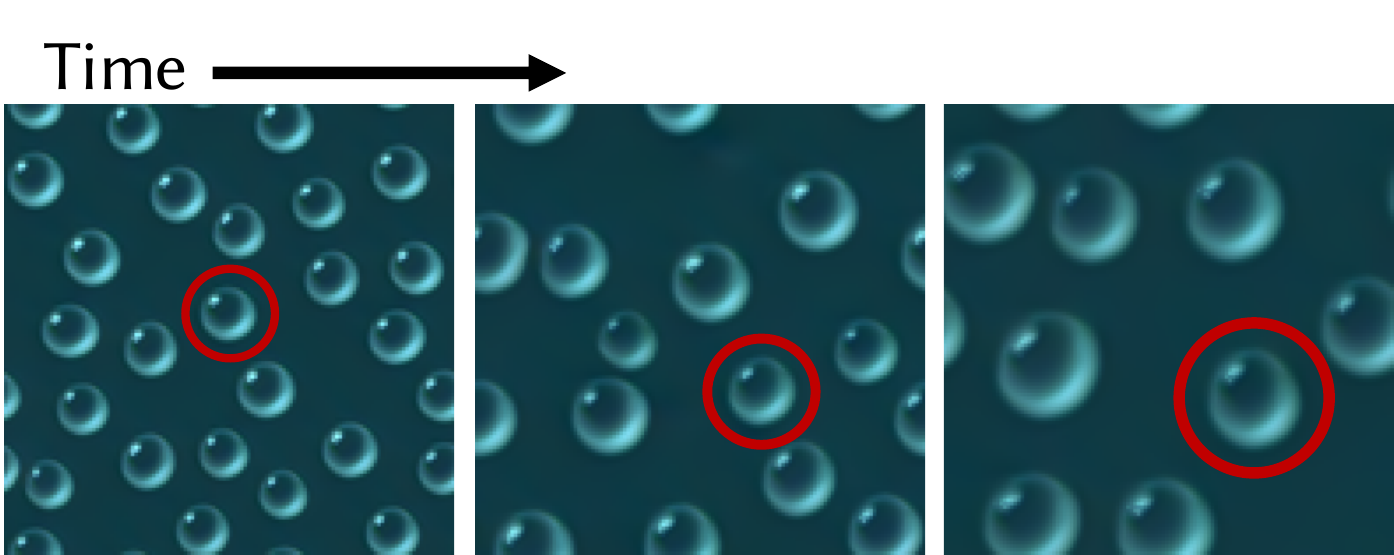}
    \caption{Varying the cell size $\Dx, \Dy$ over time $t$ causes the patterns to grow or shrink.
    The red circle indicates a bubble that grows in size while remaining coherent.}
    \label{fig:time-varying}
    \vspace{-13pt}
\end{figure}
The cell size given by the values $(\Dx, \Dy)$ in Equation~\ref{eq:pde_perception} controls the scale of the patterns. The smaller the cell size, the more cells (pixels) are required to create the pattern; therefore, the pattern becomes larger.
In this section, we demonstrate three interesting ways to control and exploit the scale of the patterns at inference time.

\subsection{Multiscale Patterns}
By setting $\Dx, \Dy$ to be functions of the cell coordinate $x, y$, we can have local control over the scale of the synthesized patterns. In this example, we use a grid of size $256\times1536$ and vary the cell size as a function of the $x$ coordinate, that is, $\Dx=\Dy=g(x)$.
Figure~\ref{fig:multiscale} shows the results for \rev{three} different textures. The target texture is shown on the left and the synthesized images demonstrating multiscale patterns are shown on the right. The function $g(x)$ is symmetric around the center of the image, increasing exponentially from $2^{-3.0}$ on the left border to $2^{0.5}$. This makes the patterns on the border to be roughly 11.0 times larger than the patterns in the center. Note how the scale of the patterns change smoothly and seamlessly across the image.

\subsection{Anisotropic Scaling}
\textit{Noise-NCA} can create anisotropically scaled patterns when $\Dx \neq \Dy$. In this example, we use two different values for each dimension of the cell size $\Dx, \Dy \in \left \{1.0, 0.5 \right \} \times \left \{1.0, 0.5 \right \}$ and set the grid size to $128 \times 128$ regardless of the cell size.
Figure~\ref{fig:anisotropic} shows the anisotropic patterns for six different textures. Decreasing $\Dx$ stretches the patterns horizontally, while decreasing $\Dy$ acts as a vertical stretch.

\vspace{-3pt}
\subsection{Time-Varying Scaling}
Varying the cell size $\Dx, \Dy$ over time $t$ causes the patterns to grow or shrink. We find that if the scale varies smoothly, the patterns remain coherent. For example, for the bubble pattern shown in Figure~\ref{fig:time-varying}, when we decrease $\Dx, \Dy$ over time, some bubbles grow coherently in size \rev{and preserve their soliton-like} structure, as indicated by the red circles.

\section{Conclusion}
In this paper, \rev{by studying the NCA behavior at the continuous space-time limit}, we investigate whether NCA learns a continuous PDE that governs its dynamics or simply overfits the discretization used during training. We show that existing NCA models struggle to generate correct textures when we change the space or time discretization, indicating overfitting. 
We analyze the source of the overfitting problem and propose \textit{Noise-NCA}, an effective and easy-to-implement solution based on using random uniform noise as the seed.
This change significantly improves NCA's ability to maintain consistent behavior across different discretization granularities and enables continuous control over the speed of the pattern formation and the scale of the synthesized patterns.

% \clearpage

\footnotesize
\bibliographystyle{apalike}
\bibliography{main} % replace by the name of your .bib file

\end{document}